\documentclass[conference]{IEEEtran}
\IEEEoverridecommandlockouts

\usepackage{cite}
\usepackage{amsmath,amssymb,amsfonts}
\usepackage{graphicx}
\usepackage{textcomp}
\usepackage{xcolor}
\usepackage{amsmath,amssymb,amsfonts}
\usepackage{graphicx}
\usepackage{textcomp}
\usepackage{xcolor}
\usepackage{amsfonts}
\usepackage{booktabs}  
\usepackage{colortbl}
\usepackage{marvosym}%
\usepackage{multirow}
\usepackage{makecell}
\usepackage{pifont} 
\usepackage{arydshln}
\usepackage{bigstrut}
\usepackage{hyperref}
\usepackage{subcaption}
\usepackage{algorithm}
\usepackage{algpseudocode}

\def\BibTeX{{\rm B\kern-.05em{\sc i\kern-.025em b}\kern-.08em
    T\kern-.1667em\lower.7ex\hbox{E}\kern-.125emX}}
\begin{document}

\title{PPGFlowECG: Latent Rectified Flow with Cross-Modal Encoding for PPG-Guided ECG Generation and Cardiovascular Disease Detection}


\author{
    \IEEEauthorblockN{
    \begin{tabular}{c}
    Xiaocheng Fang\textsuperscript{\rm1,2,3},
    Jiarui Jin\textsuperscript{\rm1,2,3},
    Haoyu Wang\textsuperscript{\rm2,5},
    Che Liu\textsuperscript{\rm4},
    Jieyi Cai\textsuperscript{\rm2,5} \\
    Yujie Xiao\textsuperscript{\rm2},
    Guangkun Nie\textsuperscript{\rm1,2,3},
    Bo Liu\textsuperscript{\rm1,3},
    Shun Huang\textsuperscript{\rm2},
    Hongyan Li\textsuperscript{\rm1,3,\textdagger},
    Shenda Hong\textsuperscript{\rm2,\textdagger}
    \end{tabular}
    }

    \IEEEauthorblockA{\textsuperscript{\rm1}School of Intelligence Science and Technology, Peking University, China}
    \IEEEauthorblockA{\textsuperscript{\rm2}National Institute of Health Data Science, Peking University, China}
    
    \IEEEauthorblockA{\textsuperscript{\rm3}State Key Laboratory of General Artificial Intelligence, Peking University, China}
    \IEEEauthorblockA{\textsuperscript{\rm4}Data Science Institute, Imperial College London, UK}
    \IEEEauthorblockA{\textsuperscript{\rm5}University of Chinese Academy of Sciences, China}
    \IEEEauthorblockA{\tt\small \{hongshenda, leehy\}@pku.edu.cn}
}

\maketitle

\begingroup
\renewcommand{\thefootnote}{}
\footnotetext{
\textsuperscript{\textdagger}Corresponding authors.
}
\endgroup

\begin{abstract}
Electrocardiography (ECG) is the clinical gold standard for cardiovascular disease (CVD) assessment, yet continuous monitoring is constrained by the need for dedicated hardware and trained personnel. Photoplethysmography (PPG) is ubiquitous in wearable devices and readily scalable, but it lacks electrophysiological specificity, limiting diagnostic reliability. While generative methods aim to translate PPG into clinically useful ECG signals, existing approaches are limited by the misalignment of physiological semantics in generative models and the complexity of modeling in high-dimensional signals. To address these limitations, we propose PPGFlowECG, a two-stage framework that aligns PPG and ECG in a shared latent space using the CardioAlign Encoder and then synthesizes ECGs with latent rectified flow. We further provide a formal analysis of this coupling, showing that the CardioAlign Encoder is necessary to guarantee stable and semantically consistent ECG synthesis under our formulation. Extensive experiments on four datasets demonstrate improved synthesis fidelity and downstream diagnostic utility. These results indicate that PPGFlowECG supports scalable, wearable-first CVD screening when standard ECG acquisition is unavailable. The code is available at \url{https://github.com/PKUDigitalHealth/PPGFlowECG}.
\end{abstract}


\section{Introduction}
Cardiovascular diseases (CVDs) remain the leading cause of mortality worldwide~\cite{timmis2022european,vaduganathan2022global,chong2025global}. Continuous monitoring can enable earlier detection and timely intervention~\cite{bayoumy2021smart}. Although electrocardiography (ECG) is the diagnostic gold standard, routine deployment is constrained by the need for specialized equipment and trained personnel~\cite{kligfield2007recommendations}. Wearable ECG devices have improved access~\cite{xie2024wearable,li2025motion}. However, short recording durations and the requirement for active user engagement still limit their suitability for long-term, passive monitoring and continuous health management~\cite{nelson2020guidelines}.

\begin{figure}[t]
\centering
\includegraphics[width=\linewidth]{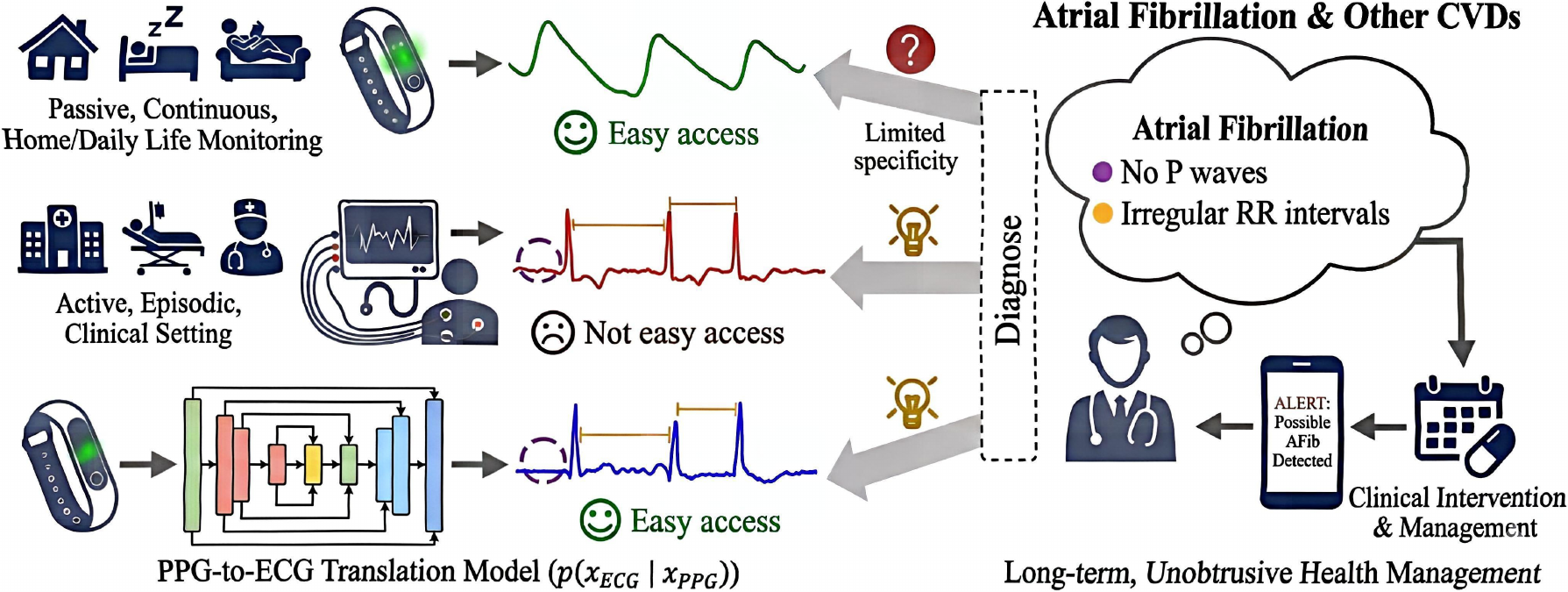}
  \caption{Motivation of PPG-to-ECG translation for wearable cardiovascular assessment. Translated ECGs provide diagnostic cues from accessible PPG signals, supporting atrial fibrillation and CVD screening.}
  \label{fig_1}
\end{figure}

Photoplethysmography (PPG) enables scalable, non-invasive, continuous monitoring and can be acquired in real time using widely deployed optical sensors. However, PPG does not directly capture cardiac electrophysiology, it provides limited diagnostic specificity for many CVDs. This creates a gap between accessible monitoring and diagnostic-grade assessment. As shown in Figure~\ref{fig_1}, PPG-to-ECG translation addresses this gap by inferring ECG-like waveforms or representations from PPG, enabling ECG-informed assessment when conventional ECG acquisition is unavailable. This is particularly relevant for long-term home monitoring, where standard ECG devices are less practical and sustained adherence is difficult. Compared with clinic-based ECG workflows that require visits and active participation, PPG-to-ECG translation supports unobtrusive, longitudinal monitoring for high-risk individuals, facilitating chronic disease management, and earlier CVD screening.

Large-scale public datasets such as MCMED~\cite{kansal2025mc}, which provide synchronized PPG–ECG recordings with curated CVD annotations, enable systematic development and evaluation of PPG-to-ECG translation methods for CVD diagnosis. Recent work has investigated diffusion models~\cite{shome2024region} and flow-based models~\cite{nambu2025cardioflow} for this task. While diffusion models often achieve strong synthesis quality, they typically require many iterative denoising steps, resulting in high inference latency. In contrast, flow-based models reduce inference cost by learning an explicit transformation from a simple noise distribution to the target signal, enabling sampling in a few steps.

Despite recent progress, PPG-to-ECG translation faces two major challenges: 1) \textbf{\textit{Physiological Semantics Misalignment in Generative Models}}: Conventional end-to-end approaches often act as mere “waveform mimickers,” focusing on low-level signal reconstruction rather than physiological semantics. As a result, generated ECGs may resemble authentic waveforms morphologically but fail to preserve clinically actionable features essential for reliable cardiovascular disease screening and diagnosis, thereby limiting practical utility. 2) \textbf{\textit{Modeling Complexity in High-Dimensional Signals}}: Complex temporal dependencies, inter-subject variability, and abrupt waveform transitions create highly non-smooth, modality-specific signal manifolds in raw data space. These properties violate the continuity and invertibility assumptions underlying conventional generative models, particularly flow-based architectures, leading to unstable training and poor generalization.

To address these challenges, we propose PPGFlowECG, a two-stage framework that couples CardioAlign Encoder with latent rectified flow for PPG-to-ECG translation and CVD detection. Stage~1 pretrains a cross-modal encoder–decoder with a shared CardioAlign encoder whose parameters are tied across PPG and ECG, encouraging modality-invariant cardiovascular representations beyond waveform-level similarity. Stage~2 trains a rectified-flow model in the latent space to learn a deterministic mapping from Gaussian noise to ECG latents via straight-line transport. Together, these stages yield ECGs with higher morphological fidelity and improved semantic consistency, facilitating clinical interpretation and downstream diagnosis. The main contributions are summarized as follows:

\begin{itemize}
\item We propose PPGFlowECG, a novel two-stage framework for high-fidelity PPG-to-ECG translation and cardiovascular disease detection. To our knowledge, this is \textit{the first latent rectified flow model} specifically designed for cross-modal physiological signal generation.

\item The proposed CardioAlign Encoder, together with multi-granularity latent constraints, induces a semantically aligned PPG–ECG latent space that couples naturally with latent rectified flow. 

\item We further provide a formal coupling analysis showing that latent alignment reduces conditional dispersion and vector-field curvature, improving numerical stability and enabling efficient few-step sampling.

\item We perform \textit{the first large-scale evaluation on MCMED}, showing consistent gains in both translation quality and downstream disease detection. Cardiologist-led studies further suggests that our method can support real-world cardiovascular screening.
\end{itemize}

\section{Related Work}

\subsection{PPG-based Cardiovascular Monitoring}

Photoplethysmography (PPG) has been widely studied as a low-cost optical sensing modality for cardiovascular monitoring in wearable and bedside settings~\cite{nie2026scoping,nie2026anyppg}. Prior studies have used PPG for wearable heart-rate measurement~\cite{bent2020investigating}, atrial-fibrillation and rhythm detection~\cite{torres2020multi,pereira2020photoplethysmography}, blood-pressure assessment~\cite{elgendi2024recommendations}, and cardiovascular digital biomarker development~\cite{nie2025artificial}. These studies establish PPG as a scalable signal source for long-term and unobtrusive cardiovascular assessment, especially when continuous ECG acquisition is impractical. PPG provides an indirect hemodynamic measurement of cardiovascular activity, whereas electrocardiography (ECG) directly records cardiac electrical activation and remains central to rhythm analysis, conduction assessment, and ECG-informed CVD screening~\cite{kligfield2007recommendations,jin2025self,jin2026ecg}. This modality gap has motivated studies that infer ECG-related information from PPG, ranging from ECG parameter estimation to waveform-level PPG-to-ECG reconstruction~\cite{banerjee2014photoecg,tian2022cross,zhu2021learning}. These works provide the foundation for PPG-guided ECG synthesis, which aims to recover diagnostically informative ECG morphology from accessible PPG signals.

\subsection{PPG-to-ECG Generation}
Early PPG-to-ECG methods mainly relied on feature-engineered or transform-based pipelines. PhotoECG estimates ECG parameters from PPG-derived features~\cite{banerjee2014photoecg}, cross-domain dictionary learning exploits shared sparse structures between PPG and ECG~\cite{tian2022cross}, and DCT-based methods approximate cross-modal mappings in a transformed signal domain~\cite{zhu2021learning}. These methods established the feasibility of ECG inference from pulse waveforms, while their handcrafted assumptions limit their ability to capture nonlinear cardiovascular dynamics, inter-subject variability, and heterogeneous raw signals.

Recent studies have shifted toward end-to-end neural reconstruction and generative modeling~\cite{fang2026ecgflowcmr,fang2026imputeecg}. Lightweight neural networks have been explored for wearable PPG-based cardiac monitoring~\cite{li2020inferring}, while PPG2ECGps~\cite{tang2023ppg2ecgps} and PPG2ECG~\cite{pinto2024inferring} reconstruct ECG waveforms using subject-specific and U-Net-based architectures. Performer~\cite{lan2023performer} further introduces transformer-based reconstruction and combines generated ECGs with native PPG for CVD detection. Beyond deterministic reconstruction, CardioGAN~\cite{sarkar2021cardiogan} and P2E-WGAN~\cite{vo2021p2e} use adversarial learning for conditional ECG synthesis, ADSSM~\cite{vo2024ppg} models latent temporal dynamics through attention-based state-space modeling, and RDDM~\cite{shome2024region} improves morphology modeling with region-disentangled diffusion. More recently, CardioFlow~\cite{nambu2025cardioflow} applies rectified flow to PPG-to-ECG translation with probabilistic peak masking to emphasize key waveform structures.

These methods have improved reconstruction fidelity, waveform realism, and downstream diagnostic utility. However, most existing approaches either optimize raw waveform similarity or learn generative dynamics directly in the signal space. Such raw-space modeling is challenging for physiological signals with abrupt QRS transitions, rhythm-dependent variations, subject-specific morphology, and modality-specific noise. This can limit semantic consistency between the input PPG and generated ECG, motivating latent-space generation with explicit cross-modal physiological alignment~\cite{fang2026echobridge}.

\begin{figure*}[t]
\centering
\includegraphics[width=0.99\textwidth]{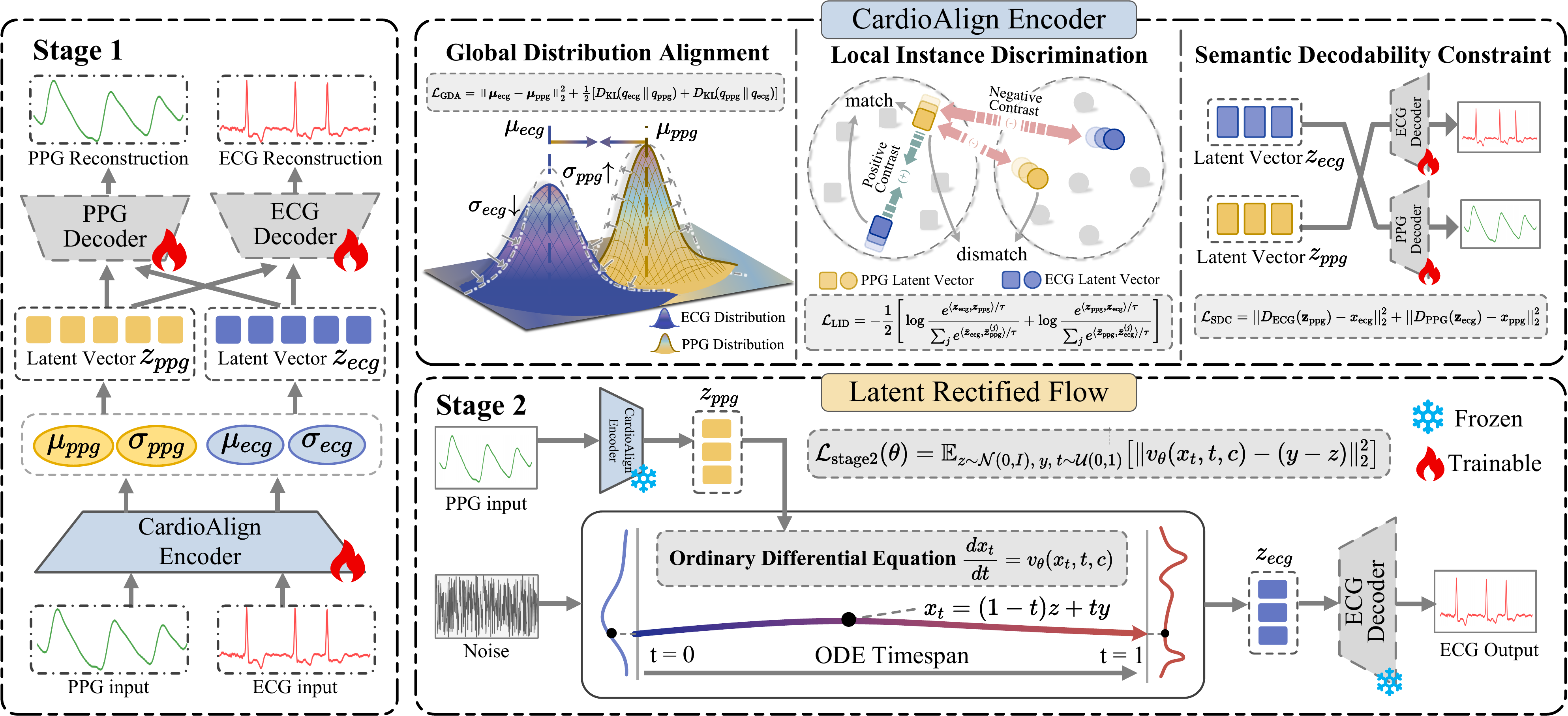}
\caption{
Overview of PPGFlowECG for PPG-guided ECG generation. 
Stage 1 learns a shared PPG--ECG latent space with global distribution alignment, local instance discrimination, and semantic decodability constraints; Stage 2 synthesizes ECG latents from PPG-conditioned latent rectified flow and decodes them into ECG waveforms.
}
\label{fig_2}
\end{figure*}

\section{Methodology}

\subsection{Overview of PPGFlowECG}
PPGFlowECG is a two-stage framework for PPG-guided ECG generation. First, the shared CardioAlign Encoder constructs a semantically aligned PPG-ECG latent space. Second, a PPG-conditioned latent rectified flow transports Gaussian noise to an ECG latent through straight-line dynamics, which is then decoded into an ECG waveform. As illustrated in Fig.~\ref{fig_2}, generation in the aligned latent space reduces the complexity of raw waveform modeling while preserving ECG morphology and disease-relevant semantics.

\subsection{CardioAlign Encoder}
PPG and ECG differ in surface morphology yet reflect the same underlying cardiovascular dynamics. When trained separately, modality-specific encoders may overfit modality-specific cues and miss shared physiological semantics. The CardioAlign Encoder mitigates this by using a shared encoder $E_{\text{CA}}(\cdot)$ for both modalities, aligning them in a unified latent space and reducing sensitivity to waveform-specific appearance while retaining clinically relevant features for downstream diagnosis. Given an input waveform $x_m \in \mathbb{R}^{L \times 1}$ from modality $m \in \{\text{ppg}, \text{ecg}\}$, the encoder outputs $\boldsymbol{\mu}_m$ and $\boldsymbol{\sigma}_m$, and samples $\mathbf{z}_m$ via the reparameterization trick:
\begin{equation}
\mathbf{z}_m = \boldsymbol{\mu}_m + \boldsymbol{\sigma}_m \odot \boldsymbol{\epsilon},\quad
\boldsymbol{\epsilon}\sim\mathcal{N}(\mathbf{0}, \mathbf{I}).
\end{equation}

To induce a semantically coherent latent space, we adopt a coarse-to-fine alignment strategy at three levels, as follows:

\subsubsection{Global Distribution Alignment}
At the distributional level, we model the latent posterior of each modality as a diagonal Gaussian. For modality $m \in \{\mathrm{ppg}, \mathrm{ecg}\}$, the posterior is defined as:
\begin{equation}
q_m =
\mathcal{N}\!\left(
\boldsymbol{\mu}_m,
\operatorname{diag}(\boldsymbol{\sigma}_m^2)
\right),
\end{equation}
where $\boldsymbol{\mu}_m$ and $\boldsymbol{\sigma}_m^2$ denote the posterior mean and variance, respectively. $\mathcal{L}_{\text{GDA}}$ is defined as follow:
\begin{equation}
\begin{aligned}
\mathcal{L}_{\text{GDA}}
&= \left\lVert \boldsymbol{\mu}_{\text{ecg}}-\boldsymbol{\mu}_{\text{ppg}} \right\rVert_2^2 \\
&\quad + \tfrac{1}{2}\!\left[
D_{\text{KL}}\!\left(q_{\text{ecg}} \,\|\, q_{\text{ppg}}\right)
+ D_{\text{KL}}\!\left(q_{\text{ppg}} \,\|\, q_{\text{ecg}}\right)
\right].
\end{aligned}
\end{equation}

The first term aligns the posterior means, while the symmetric Kullback--Leibler (KL) divergence~\cite{zhang2023properties} matches posterior dispersion across modalities. Together, these terms impose a stable distribution-level constraint for learning a shared physiological latent space.

\subsubsection{Local Instance Discrimination} While global distribution alignment provides a stable coarse constraint across modalities, it lacks the granularity needed to preserve patient-specific cardiac signatures and rhythm-level variations. We therefore propose a Local Instance Discrimination (LID) mechanism. Let $\bar{\mathbf{z}}_{\text{ecg}}$ and $\bar{\mathbf{z}}_{\text{ppg}}$ denote mean-pooled latent representations. Paired segments are treated as positives, whereas cross-subject in-batch pairings serve as negatives, encouraging $E_{\text{CA}}(\cdot)$ to align the two modalities while retaining subject-specific information. $\mathcal{L}_{\text{LID}}$ is defined as follow:
\begin{equation}
\mathcal{L}_{\text{LID}}
= -\frac{1}{2} \Bigg[\log
\frac{
e^{\langle \bar{\boldsymbol z}_{\text{ecg}},
        \bar{\boldsymbol z}_{\text{ppg}} \rangle / \tau}
}{
\sum_{j}
e^{\langle \bar{\boldsymbol z}_{\text{ecg}},
        \bar{\boldsymbol z}_{\text{ppg}}^{(j)} \rangle / \tau}
}
+
\log
\frac{
e^{\langle \bar{\boldsymbol z}_{\text{ppg}},
        \bar{\boldsymbol z}_{\text{ecg}} \rangle / \tau}
}{
\sum_{j}
e^{\langle \bar{\boldsymbol z}_{\text{ppg}},
        \bar{\boldsymbol z}_{\text{ecg}}^{(j)} \rangle / \tau}
}
\Bigg].
\end{equation}

By increasing similarity for matched cross-modal pairs and decreasing similarity for mismatched subjects, $\mathcal{L}_{\text{LID}}$ mitigates representation over-smoothing in the aligned latent space and promotes subject-discriminative embeddings.

\subsubsection{Semantic Decodability Constraint} While the objectives above establish coarse-to-fine cross-modal alignment, a unified representation should not only reside in a shared latent space but also support consistent cross-modal decoding. To impose this stronger functional constraint, we require the latent representation from one modality to reconstruct the other. This cross-modal decodability encourages the latent space to encode translatable physiological factors necessary for modality fusion. We define $\mathcal{L}_{\text{SDC}}$ as follows:
\begin{equation}
\mathcal{L}_{\text{SDC}} = ||D_{\text{ECG}}(\mathbf{z}_{\text{ppg}}) - x_{\text{ecg}}||_2^2
+ ||D_{\text{PPG}}(\mathbf{z}_{\text{ecg}}) - x_{\text{ppg}}||_2^2.
\label{eq:cross_loss}
\end{equation}

By enforcing $\mathbf{z}_{\text{ppg}}$ to decode into ECG and $\mathbf{z}_{\text{ecg}}$ into PPG, the latent space is driven to capture physiologically meaningful factors that generalize across modalities, thereby reinforcing cross-modal translatability. Overall, $\mathcal{L}_{\text{CA}}$ is formulated as:
\begin{equation}
\mathcal{L}_{\text{CA}}
= \mathcal{L}_{\text{GDA}}
+ \mathcal{L}_{\text{LID}}
+ \mathcal{L}_{\text{SDC}}.
\end{equation}

\subsection{Latent Rectified Flow}
To enable high-fidelity ECG generation from PPG, we develop a generative model based on latent rectified flow. This formulation reduces generation to learning a straight-line vector field in a structured latent space. Rather than modeling flow dynamics directly in the raw waveform domain~\cite{nambu2025cardioflow}, we operate in the shared latent space, which is lower-dimensional and smoother. This choice better satisfies the continuity assumptions underlying flow models, improving training stability and enabling efficient sampling.

\subsubsection{Conditional Generative Modeling}
Building on the aligned latent manifold, we employ latent rectified flow to synthesize high-fidelity ECG signals. Specifically, we learn a conditional vector field $v_{\theta}$ that transports samples from an isotropic Gaussian prior $z \sim \mathcal{N}(\mathbf{0}, \mathbf{I})$ to the target ECG latent representation $y$. For interpolation time $t \in [0,1]$, we define the probability-flow dynamics as follows:
\begin{equation}
x_t = (1 - t)z + ty.
\end{equation}

Here, $t=0$ and $t=1$ correspond to the noise prior and the target latent, respectively. Under linear interpolation, the optimal drift (instantaneous velocity) is constant, i.e., $v^{*}(x_t,t,c)=y-z$. The CardioAlign Encoder concentrates the conditional distribution $p(y\mid c)$, which in turn reduces the curvature of the learned vector field $v_{\theta}(x_t,t,c)$. We optimize $v_{\theta}$ using a mean-squared error objective:
\begin{equation}
\mathcal{L}(\theta)=\mathbb{E}_{z\sim\mathcal{N}(0,I),y,t\sim\mathcal{U}(0,1)}\left[\|v_{\theta}(x_{t},t,c)-(y-z)\|_{2}^{2}\right].
\end{equation}

To estimate $v_{\theta}$, we use a Transformer backbone with hierarchical cross-modal conditioning. The ECG latent $y$ is projected into a compact token sequence via a 1D convolutional embedder, preserving local temporal structure while supporting long-range modeling. The PPG latent $c$ conditions generation through cross-attention in every Transformer block, with $c$ as key-value and the evolving ECG tokens as queries. This layerwise conditioning promotes fine-grained temporal alignment and physiological consistency along the generative trajectory, enabling clinically interpretable ECG synthesis with few ODE-solver steps.  

\subsubsection{Inference and Sampling} During inference, we synthesize an ECG signal conditioned on the PPG latent $c$. We first sample an initial noise vector $z \sim \mathcal{N}(\mathbf{0}, \mathbf{I})$ as the starting point of the generative trajectory. The mapping from $z$ to the target ECG latent $y$ is then defined by an ODE~\cite{hartman2002ordinary} governed by the learned conditional vector field $v_{\theta}$:
\begin{equation}
    \frac{d x_t}{dt} = v_\theta(x_t, t, c), 
    \quad t \in [0, 1].
\end{equation}

Here, $x_t$ denotes the latent state at time $t$. To approximate the continuous trajectory, we use an explicit Euler solver with a fixed number of steps $T$. Let $\Delta t = 1/T$ and $t_k = k\Delta t$, the latent state is updated iteratively as follows:
\begin{equation}
\begin{aligned}
x_{k+1}
&= x_k + \Delta t \cdot v_\theta(x_k, t_k, c),
\quad k = 0, 1, \dots, T-1.
\end{aligned}
\end{equation}

Finally, the terminal state $x_T$ is taken as the generated ECG latent representation. This latent is then decoded by the frozen ECG decoder to reconstruct the synthesized ECG.

\subsection{Analysis of PPGFlowECG}
The effectiveness of PPGFlowECG comes from coupling cross-modal latent alignment with rectified-flow transport. We analyze this coupling from two perspectives: how CardioAlign reduces the ambiguity of conditional ECG generation, and how the aligned latent space promotes straighter trajectories for stable few-step sampling.

\subsubsection{Conditional Dispersion and Drift Ambiguity}
Let $c=z_{\mathrm{ppg}}$ and $y=z_{\mathrm{ecg}}$ denote the conditioning and target latents in the shared space. Stage~1 optimizes the $\mathcal{L}_{\mathrm{CA}}$ to align paired PPG and ECG latents. We measure the remaining target uncertainty by the average conditional dispersion:
\begin{equation}
\bar{\kappa}
=
\mathbb{E}_{c}
\left[
\operatorname{tr}\left(
\operatorname{Cov}(y\mid c)
\right)\right].
\end{equation}

A smaller $\bar{\kappa}$ indicates that ECG latents are more concentrated under each PPG condition, reducing target ambiguity for PPG-conditioned generation.

In Stage~2, rectified flow estimates the straight-line drift $y-z$ under a mean-squared objective. The Bayes uncertainty of this drift regression can be as follows:
\begin{equation}
\mathcal{R}_{\mathrm{drift}}
=
\mathbb{E}_{x_t,t,c}
\left[
\operatorname{tr}\left(
\operatorname{Cov}
\left(
y-z
\,\middle|\,
x_t,t,c
\right)\right)
\right].
\end{equation}

This term measures the ambiguity of the target velocity under the same interpolation state and PPG condition. Since the uncertainty of $y$ under $c$ is a major source of drift ambiguity, reducing $\bar{\kappa}$ through $\mathcal{L}_{\mathrm{CA}}$ makes the ECG target latent more concentrated for each PPG latent. This improves the conditioning of the vector-field regression and helps preserve fine-grained ECG morphology.

\begin{table*}[t]
\centering
\footnotesize
\setlength{\tabcolsep}{3.5pt}
\renewcommand{\arraystretch}{1.05}
\caption{Comparison with state-of-the-art approaches on four datasets. Lower values are better for all metrics. Best and second-best results are highlighted in bold and underline, respectively.}
\label{Table1}
\begin{tabular}{@{}l c *{12}{c}@{}}
\toprule
\multirow{2}{*}{\textbf{Method}} 
& \multirow{2}{*}{\textbf{Ref.}}
& \multicolumn{3}{c}{\textbf{MCMED}} 
& \multicolumn{3}{c}{\textbf{MIMIC-AFib}} 
& \multicolumn{3}{c}{\textbf{VitalDB}} 
& \multicolumn{3}{c}{\textbf{BIDMC}} \\
\cmidrule(lr){3-5}
\cmidrule(lr){6-8}
\cmidrule(lr){9-11}
\cmidrule(l){12-14}
& 
& \textbf{MAE} 
& \textbf{RMSE} 
& $\boldsymbol{\mathrm{MAE}_{\mathrm{HR}}}$ 
& \textbf{MAE} 
& \textbf{RMSE} 
& $\boldsymbol{\mathrm{MAE}_{\mathrm{HR}}}$ 
& \textbf{MAE} 
& \textbf{RMSE} 
& $\boldsymbol{\mathrm{MAE}_{\mathrm{HR}}}$ 
& \textbf{MAE} 
& \textbf{RMSE} 
& $\boldsymbol{\mathrm{MAE}_{\mathrm{HR}}}$ \\
\midrule
DDPM~\cite{ho2020denoising} 
& NeurIPS'20
& \underline{0.94} & 1.36 & 7.00 
& \textbf{0.73} & \textbf{1.22} & 9.55 
& 0.83 & \underline{1.31} & 11.57 
& \underline{0.79} & \underline{1.24} & 5.82 \\

CardioGAN~\cite{sarkar2021cardiogan} 
& AAAI'21
& 0.98 & 1.40 & \underline{2.72} 
& \textbf{0.73} & \textbf{1.22} & \underline{3.42} 
& \underline{0.82} & 1.32 & \underline{3.50} 
& 0.82 & 1.31 & \textbf{1.38} \\

Rectified Flow~\cite{liuflow} 
& ICLR'23
& 0.97 & 1.36 & 26.40 
& 0.83 & 1.36 & 26.01 
& 0.90 & 1.37 & 36.36 
& 0.82 & 1.33 & 9.14 \\

RDDM~\cite{shome2024region} 
& AAAI'24
& 0.99 & 1.41 & 7.45 
& 0.82 & 1.37 & 15.35 
& 0.87 & 1.40 & 30.27 
& 0.83 & 1.34 & 10.23 \\

CardioFlow~\cite{nambu2025cardioflow} 
& ICASSP'25
& 0.95 & \underline{1.35} & 15.37 
& 0.76 & 1.33 & 17.92 
& 0.85 & 1.34 & 28.87 
& \underline{0.79} & 1.31 & 7.37 \\

\midrule
\textbf{PPGFlowECG} 
& Ours
& \textbf{0.73} & \textbf{1.14} & \textbf{1.80} 
& \textbf{0.73} & \underline{1.29} & \textbf{2.30} 
& \textbf{0.59} & \textbf{1.10} & \textbf{3.23} 
& \textbf{0.71} & \textbf{1.15} & \underline{2.35} \\
\bottomrule
\end{tabular}
\end{table*}

\subsubsection{Trajectory Linearization and Numerical Stability}
CardioAlign Encoder also improves sampling by encouraging straighter latent transport trajectories. In the ideal aligned case where the ECG target latent is deterministic under the PPG condition, i.e., $y=g(c)$, the rectified-flow target velocity along the interpolation path becomes the constant displacement $v^{\star}(x_t,t,c)=y-z=g(c)-z$, and thus $\frac{d}{dt}v^{\star}(x_t,t,c)=0$. This indicates that a well-aligned latent space turns PPG-conditioned ECG generation into a near-linear transport problem. For the learned vector field, we measure trajectory curvature by:
\begin{equation}
\mathcal{C}
=
\mathbb{E}_{x_t,t,c}
\left[
\left\|
\frac{d}{dt}v_{\theta}(x_t,t,c)
\right\|_2^2
\right].
\end{equation}
With an explicit Euler solver of step size $\Delta t$, the local truncation error scales as:
\begin{equation}
\|e_k\|
=
\mathcal{O}
\left(
\Delta t^2
\sup_{t\in[t_k,t_{k+1}]}
\left\|
\frac{d}{dt}v_{\theta}(x_t,t,c)
\right\|_2
\right).
\end{equation}

Thus, lower-curvature trajectories reduce numerical error during ODE sampling. By reducing conditional target ambiguity, CardioAlign Encoder encourages such trajectories, enabling Stage~2 to perform stable and efficient latent transport.

\begin{table}[t]
\centering
\caption{Dataset statistics after preprocessing. The last three columns report the numbers of high-quality 10-second PPG--ECG segment pairs.}
\label{tab:dataset_statistics}

\footnotesize
\setlength{\tabcolsep}{2.5pt}
\renewcommand{\arraystretch}{1.05}

\begin{tabular}{@{}lccrrr@{}}
\toprule
\textbf{Dataset} &
\textbf{Patients/Visits} &
\textbf{Label} &
\multicolumn{3}{c}{\textbf{10-s Segments}} \\
\cmidrule(lr){4-6}
& & &
\textbf{Train} &
\textbf{Val.} &
\textbf{Test} \\
\midrule
MCMED~\cite{kansal2025mc}
& 118,385
& CVD
& 10,356,840
& 1,299,869
& 37,052 \\

MIMIC-AFib~\cite{bashar2019noise}
& 35
& AF
& 2,044
& --
& 577 \\

VitalDB~\cite{lee2022vitaldb}
& 6,388
& --
& 1,007,165
& --
& 10,720 \\

BIDMC~\cite{pimentel2016toward}
& 53
& --
& 1,458
& --
& 326 \\
\bottomrule
\end{tabular}

\end{table}



\begin{figure}[t]
    \centering
    \begin{subfigure}{0.43\linewidth}
        \centering
        \includegraphics[width=\columnwidth]{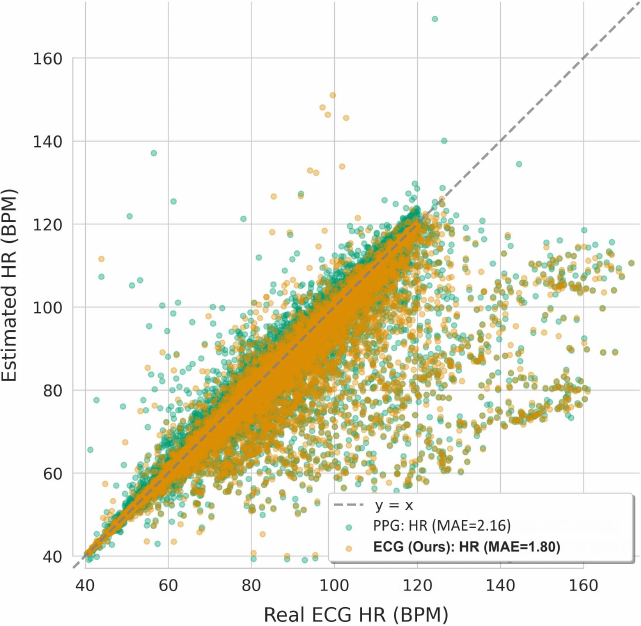}
        \caption{}
        \label{Fig3a}
    \end{subfigure}
    \begin{subfigure}{0.49\linewidth}
        \centering
        \includegraphics[width=\columnwidth]{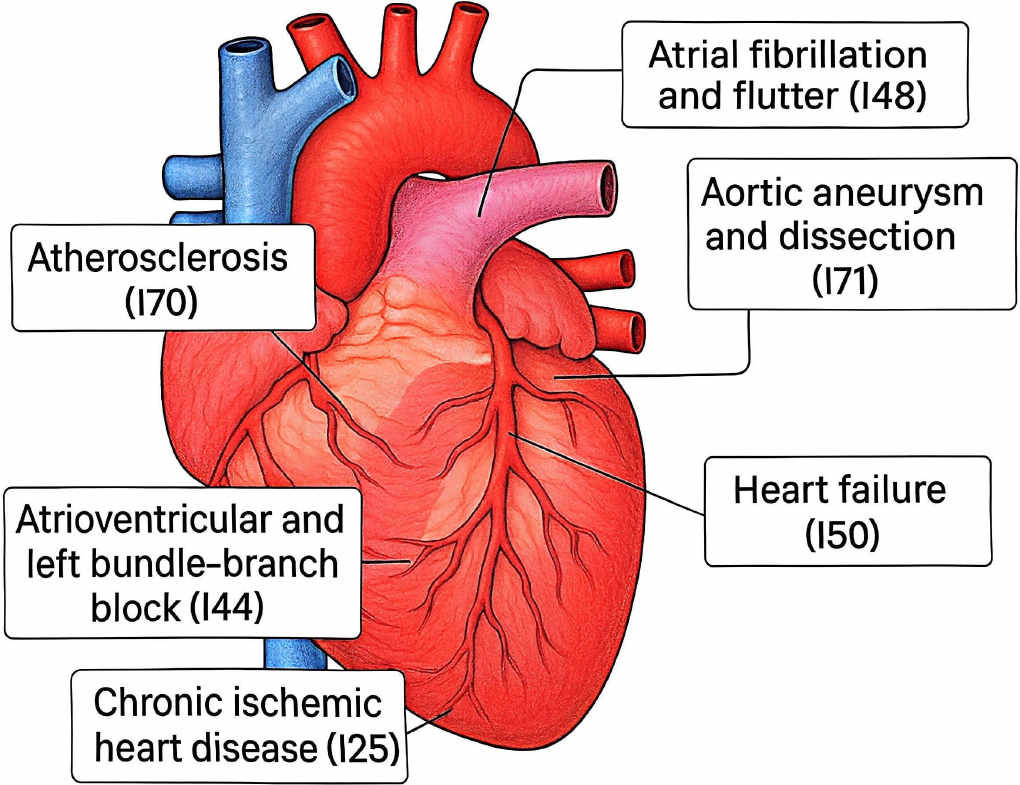}
        \caption{}
        \label{Fig3b}
    \end{subfigure}
    \caption{(a) Heart-rate estimation consistency. PPGFlowECG-generated ECGs yield HR estimates closer to real ECGs than raw PPG. (b) Downstream CVD categories evaluated on MCMED, spanning rhythm, conduction, vascular, and structural disorders.}
    \label{Fig3}
\end{figure}

\section{Experiments}
\subsection{Datasets and Evaluation Metrics}
We evaluate PPGFlowECG on four datasets spanning diverse clinical scenarios: MCMED~\cite{kansal2025mc}, MIMIC-AFib~\cite{bashar2019noise}, VitalDB~\cite{lee2022vitaldb}, and BIDMC~\cite{pimentel2016toward}. Dataset statistics are summarized in Table~\ref{tab:dataset_statistics}. Following prior work~\cite{shome2024region}, Lead~II ECG is used as the reference signal across all datasets. For waveform fidelity, we report mean absolute error (MAE) and root mean square error (RMSE). For heart-rate consistency, we estimate heart rate from real and synthesized ECGs using the Hamilton algorithm~\cite{hamilton2002open} and report $\mathrm{MAE}_{\mathrm{HR}}$. For diagnostic utility, we report per-class AUROC and Macro-AUROC on MCMED, and accuracy and F1 score for binary atrial fibrillation classification on MIMIC-AFib.

\subsection{Implementation Details and Baselines}
ECG and PPG signals were preprocessed using standardized procedures to ensure data quality and cross-dataset consistency~\cite{pan1985real}. Continuous recordings were segmented into non-overlapping 10-s windows. All signals were resampled to 128~Hz and z-score normalized to mitigate differences in sampling rates and inter-subject variability. MCMED followed its official data split, whereas the remaining datasets were partitioned at the subject level into 80\% training and 20\% testing sets to prevent information leakage. The MCMED and VitalDB evaluations used subsets of their respective test splits. We compare PPGFlowECG with representative GAN-, diffusion-, and flow-based generative baselines. All experiments were conducted on a single NVIDIA A800 GPU.

\begin{table*}[t]
\centering
\small
\setlength{\tabcolsep}{12.0pt}
\renewcommand{\arraystretch}{1.05}
\caption{Downstream multi-label CVD classification on MCMED. 
AUROC (\%) is reported for each ICD category and Macro-AUROC; best and second-best results are bolded and underlined.}
\label{tab:mcmend_cvd_classification}
\begin{tabular}{@{}lcccccccc@{}}
\toprule
\textbf{Method} & \textbf{Ref.} 
& \textbf{I48} & \textbf{I71} & \textbf{I70} 
& \textbf{I44} & \textbf{I25} & \textbf{I50} & \textbf{Macro-AUROC} \\
\midrule
\multicolumn{9}{@{}l}{\textit{PPG-based}} \\
PaPaGei~\cite{pillaipapagei}
& ICLR'25
& 61.0 & 57.3 & 55.1
& 58.9 & 55.7 & 55.4 & 57.2 \\

\midrule
\multicolumn{9}{@{}l}{\textit{Gen.ECG-based}} \\
DDPM~\cite{ho2020denoising} 
& NeurIPS'20 
& \underline{66.6} & 60.5 & \underline{57.6} 
& \textbf{62.5} & \underline{58.4} & \underline{59.1} & \underline{60.8} \\

CardioGAN~\cite{sarkar2021cardiogan} 
& AAAI'21 
& 61.2 & \textbf{65.7} & 56.7 
& 57.0 & 57.4 & 53.9 & 58.7 \\

Rectified Flow~\cite{liuflow} 
& ICLR'23 
& 49.9 & 50.2 & 50.4 
& 49.6 & 49.7 & 49.7 & 49.9 \\

RDDM~\cite{shome2024region} 
& AAAI'24 
& 57.4 & 57.6 & 53.8 
& 57.3 & 54.4 & 52.5 & 55.5 \\

CardioFlow~\cite{nambu2025cardioflow} 
& ICASSP'25 
& 55.3 & 53.4 & 52.1 
& 54.6 & 52.6 & 51.8 & 53.3 \\

\midrule
\textbf{PPGFlowECG} 
& Ours
& \textbf{70.8} & \underline{62.6} & \textbf{62.2} 
& \underline{61.9} & \textbf{60.8} & \textbf{60.4} & \textbf{63.1} \\
\bottomrule
\end{tabular}
\end{table*}

\begin{table}[t]
\centering
\small
\setlength{\tabcolsep}{9.5pt}
\renewcommand{\arraystretch}{1.05}
\caption{AF detection performance on the MIMIC-AFib dataset. Higher values are better.}
\label{tab:mimic_afib_classification}
\begin{tabular}{@{}lccc@{}}
\toprule
\textbf{Method} & \textbf{Ref.} & \textbf{Acc.}$\uparrow$ & \textbf{F1}$\uparrow$ \\
\midrule
\multicolumn{4}{@{}l}{\textit{PPG-based}} \\
Shen et al.~\cite{shen2019ambulatory}
& KDD'19 & 0.54 & 0.51 \\

PaPaGei~\cite{pillaipapagei}
& ICLR'25 & 0.69 & 0.75 \\

\midrule
\multicolumn{4}{@{}l}{\textit{Gen.ECG-based}} \\
DDPM~\cite{ho2020denoising} 
& NeurIPS'20 & 0.70 & 0.80 \\

CardioGAN~\cite{sarkar2021cardiogan} 
& AAAI'21 & 0.71 & 0.83 \\

Rectified Flow~\cite{liuflow} 
& ICLR'23 & 0.61 & 0.71 \\

RDDM~\cite{shome2024region} 
& AAAI'24 & 0.70 & 0.82 \\

CardioFlow~\cite{nambu2025cardioflow} 
& ICASSP'25 & 0.68 & 0.77 \\

\midrule
\textbf{PPGFlowECG (Internal)} 
& Ours & \textbf{0.82} & \textbf{0.87} \\

\textbf{PPGFlowECG (External)} 
& Ours & 0.77 & 0.82 \\
\bottomrule
\end{tabular}
\end{table}

\subsection{Comparisons with State-of-The-Arts}

\subsubsection{Evaluation on PPG-to-ECG Translation}

Table~\ref{Table1} presents the quantitative comparison across four datasets. PPGFlowECG achieves the best performance on 10 of the 12 evaluated metrics and ranks second on the remaining two. On MCMED, it consistently outperforms the strongest baseline, reducing MAE from 0.94 to 0.73, RMSE from 1.35 to 1.14, and $\mathrm{MAE}_{\mathrm{HR}}$ from 2.72 to 1.80. On VitalDB, PPGFlowECG further reduces MAE and RMSE from 0.82 and 1.31 to 0.59 and 1.10, respectively. On MIMIC-AFib, it achieves the lowest $\mathrm{MAE}_{\mathrm{HR}}$ while matching the best MAE. On BIDMC, it obtains the lowest MAE and RMSE and the second-lowest $\mathrm{MAE}_{\mathrm{HR}}$.

Fig.~\ref{Fig3a} further assesses heart-rate consistency. HR estimates derived from ECGs generated by PPGFlowECG are more tightly concentrated around the $y=x$ identity line than those derived from PPG signals. Quantitatively, our method reduces the HR estimation error from 2.16 to 1.80, corresponding to a relative reduction of 16.7\%. These results demonstrate that PPGFlowECG effectively preserves waveform fidelity and physiologically meaningful rhythm information.

\subsubsection{Evaluation on Cardiovascular Disease Detection}

We further evaluate whether ECGs generated by PPGFlowECG retain disease-discriminative information for downstream cardiovascular disease detection on MCMED and MIMIC-AFib. For MCMED, we conduct multi-label cardiovascular classification over ICD-defined categories covering diverse pathophysiological mechanisms, including rhythm, conduction, vascular, and structural disorders, as shown in Fig.~\ref{Fig3b}. To ensure a controlled comparison, all methods are evaluated using the same Net-1D classifier~\cite{hong2020holmes}. Table~\ref{tab:mcmend_cvd_classification} shows that PPGFlowECG achieves the highest Macro-AUROC of 63.1, outperforming the PPG-based baseline PaPaGei by 5.9 AUROC points and the strongest Gen.ECG-based baseline by 2.3 AUROC points. It also ranks first in four of the six disease categories and second in the remaining two, indicating consistent improvements across heterogeneous cardiovascular conditions.

For MIMIC-AFib, we evaluate binary atrial fibrillation detection using a VGG-13 classifier~\cite{sallem2020detection}. As reported in Table~\ref{tab:mimic_afib_classification}, PPGFlowECG achieves an accuracy of 0.82 and an F1 score of 0.87 in the internal setting, improving over the strongest PPG-based baseline by 0.13 and 0.12 in accuracy and F1, respectively, and over the strongest Gen.ECG-based baseline by 0.11 and 0.04. These results demonstrate that ECGs generated by PPGFlowECG preserve diagnostically relevant information for downstream cardiovascular prediction.

\begin{table}[t]
\centering
\small
\setlength{\tabcolsep}{8pt}
\renewcommand{\arraystretch}{1.08}
\caption{Ablation study on the MCMED dataset.}
\label{tab:ablation_variants}
\begin{tabular}{@{}lcccc@{}}
\toprule
\textbf{Variant} 
& \textbf{CAE} 
& \textbf{LRF} 
& \textbf{$\text{MAE}_{\text{HR}}$} 
& \textbf{Macro-AUROC} \\
\midrule
Baseline 
& -- 
& -- 
& 3.28 
& 53.1 \\

w/ CAE 
& \checkmark 
& -- 
& 1.93 
& 61.5 \\

w/ LRF 
& -- 
& \checkmark 
& 2.12 
& 58.9 \\

Full model 
& \checkmark 
& \checkmark 
& \textbf{1.80} 
& \textbf{63.1} \\
\bottomrule
\end{tabular}
\end{table}

\begin{figure*}[t]
\centering
\includegraphics[width=0.99\linewidth]{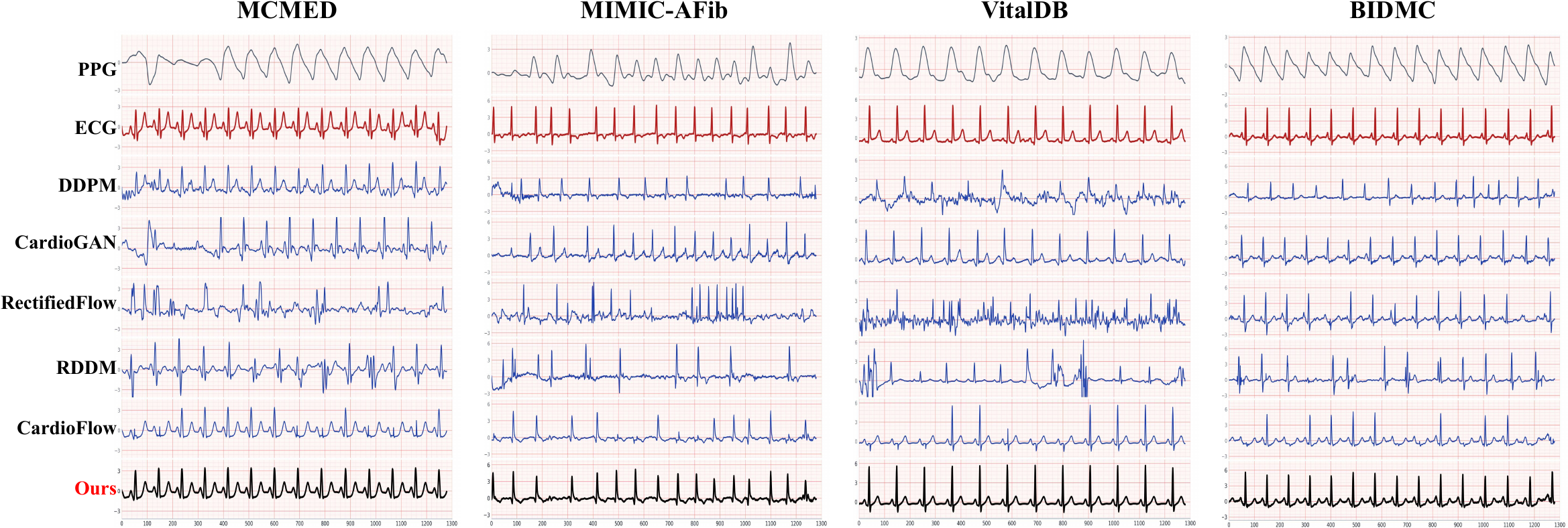}
\caption{Qualitative comparison of PPG-guided ECG synthesis across four datasets. Rows show the input PPG, reference ECG, and ECGs generated by different methods.}
\label{fig_4}
\end{figure*}

\begin{figure}[t]
\centering
\includegraphics[width=0.99\linewidth]{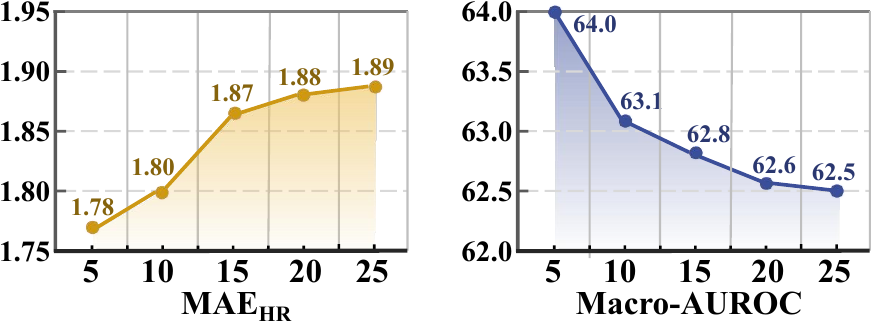}
\caption{Ablation study of the effect of sampling steps.}
\label{fig_app2}
\end{figure}

\subsection{Ablation Study}
We evaluate four model variants to disentangle the contributions of the CardioAlign Encoder and latent rectified flow. The baseline removes both modules and uses a direct PPG-to-ECG encoder-decoder trained with waveform reconstruction loss. The w/ CardioAlign Encoder variant introduces the shared CardioAlign Encoder. At inference, the aligned PPG latent is directly decoded into ECG signals. The w/ Latent Rectified Flow variant uses separate PPG and ECG encoders and applies latent rectified flow from the conditional PPG latent to the target ECG latent. The full model integrates both modules and performs PPG-conditioned straight-line transport in the semantically aligned PPG-ECG latent space.

As reported in Table~\ref{tab:ablation_variants}, the baseline obtains an $\mathrm{MAE}_{\mathrm{HR}}$ of 3.28 and a Macro-AUROC of 53.1. The w/ CardioAlign Encoder variant improves these scores to 1.93 and 61.5, respectively, demonstrating the importance of cross-modal semantic alignment. The w/ Latent Rectified Flow variant achieves an $\mathrm{MAE}_{\mathrm{HR}}$ of 2.12 and a Macro-AUROC of 58.9, showing the benefit of latent-space generative transport. The full model achieves the best performance, with an $\mathrm{MAE}_{\mathrm{HR}}$ of 1.80 and a Macro-AUROC of 63.1. These results indicate that CardioAlign constructs a semantically aligned latent space for PPG--ECG translation, while latent rectified flow further improves ECG synthesis through stable conditional transport. Fig.~\ref{fig:supp_tsne} in the supplement visualizes PPG and ECG representations from the modality-specific and shared encoders.

\subsection{Parameter Analysis}
We analyze the effect of the number of ODE solver steps by varying $T$ from 5 to 25. As shown in Fig.~\ref{fig_app2}, PPGFlowECG achieves the best performance at $T=5$, while larger step counts lead to a gradual performance decline. For the main experiments, we set $T=10$ to align with the default inference protocol of baseline methods such as RDDM and ensure comparable evaluation conditions. Under this matched setting, PPGFlowECG maintains strong performance across the reported metrics.

\begin{figure}[t]
\centering
\includegraphics[width=0.95\linewidth]{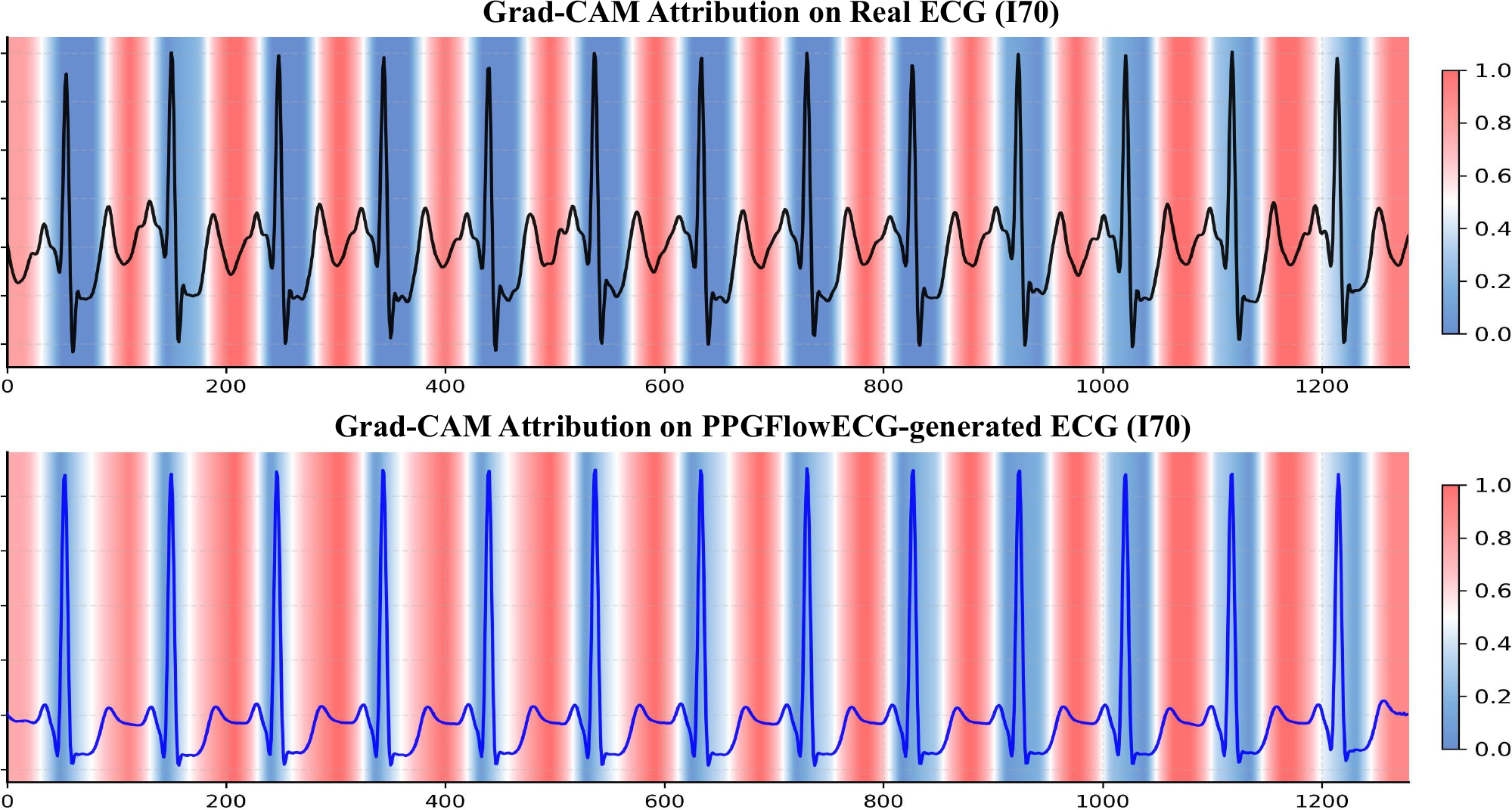}
\caption{Grad-CAM attribution maps for I70 diagnosis on real and PPGFlowECG-generated ECGs.}
\label{fig_15}
\end{figure}

\subsection{Generalization Analysis}
To assess the zero-shot generalization of PPGFlowECG under cross-dataset distribution shift, we compare two validation settings: internal validation, where the model is trained and evaluated on MIMIC-AFib, and external validation, where the model is trained on MCMED and directly evaluated on MIMIC-AFib in a zero-shot manner. As reported in Table~\ref{tab:mimic_afib_classification}, PPGFlowECG achieves an accuracy of 0.82 and an F1 score of 0.87 in the internal setting, and an accuracy of 0.77 and an F1 score of 0.82 in the external setting. The limited performance gap indicates robust cross-dataset transfer and supports the practical applicability of PPGFlowECG when paired target-domain PPG-ECG data are scarce.

\subsection{Visualization Results}

\subsubsection{Comparison of Synthesized ECGs}
We qualitatively compare ECG waveforms synthesized by different methods to assess generation quality. As shown in Fig.~\ref{fig_4}, PPGFlowECG more faithfully preserves clinically relevant morphology, producing sharper and better-aligned QRS complexes, more plausible ST--T dynamics, and fewer spurious oscillations. Overall, these examples indicate improved temporal coherence and closer resemblance to real ECG waveforms.

\subsubsection{Explainability Analysis with Grad-CAM}
We further employ Grad-CAM~\cite{selvaraju2017grad} to examine whether synthesized ECGs elicit classifier attributions over diagnostically relevant waveform regions. Using the same classifier, we compare attribution maps between real ECGs and PPGFlowECG-generated ECGs. As shown in Fig.~\ref{fig_15}, the maps exhibit partial correspondence, although some discrepancies remain. Overall, these results suggest that the synthesized ECGs retain task-relevant features used by the downstream classifier.

\begin{figure}[t]
\centering
\includegraphics[width=0.99\linewidth]{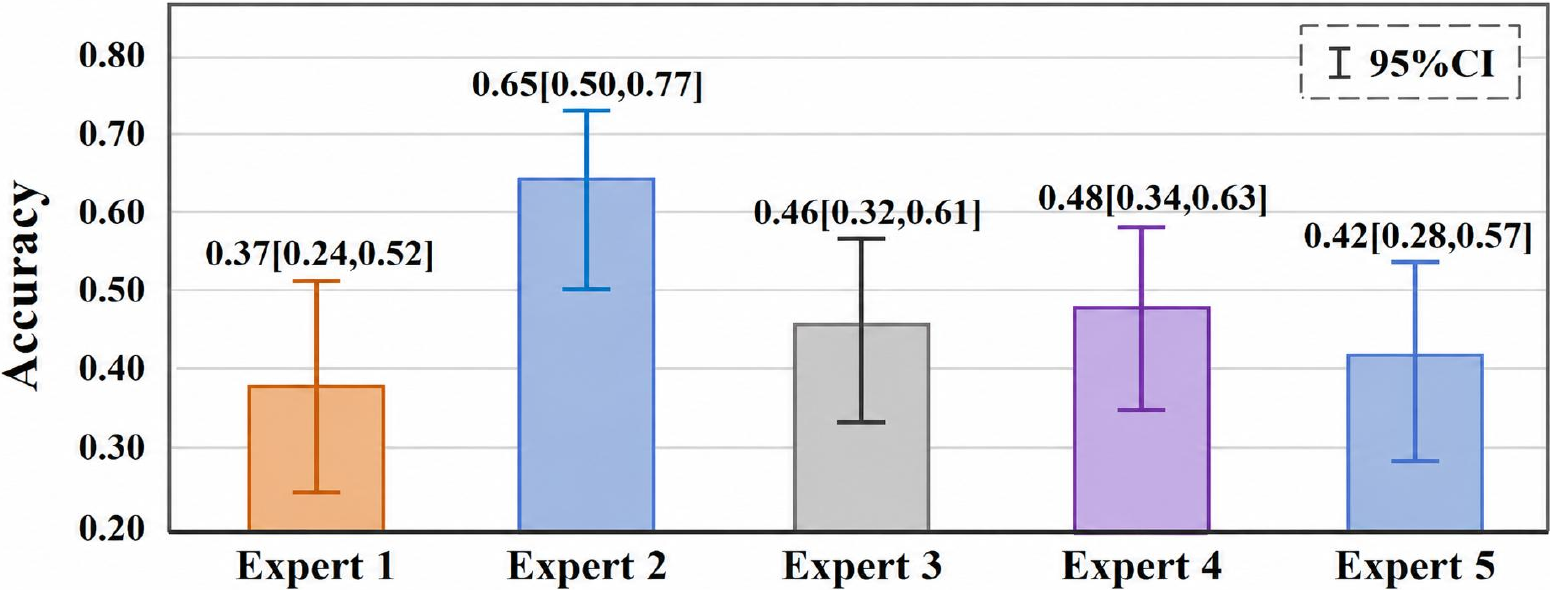}
\caption{Performance of five cardiologists on the Turing Test.}
\label{table_turing_test_stat}
\end{figure}

\subsection{Clinical Evaluation and Expert Assessment}
\subsubsection{Clinical Turing Test}

In the first task, five board-certified cardiologists, comprising one junior, three mid-level, and one senior expert, independently distinguished synthesized ECGs from real ECGs using a randomized set of 100 signals, including 50 real and 50 synthesized samples. The 50 underlying cases were distributed across six ICD-defined cardiovascular categories: atrial fibrillation and flutter (I48, $n=9$), atrioventricular and left bundle-branch block (I44, $n=9$), chronic ischemic heart disease (I25, $n=8$), heart failure (I50, $n=8$), atherosclerosis (I70, $n=8$), and aortic aneurysm and dissection (I71, $n=8$). As shown in Fig.~\ref{table_turing_test_stat}, most experts achieved identification accuracies close to chance level ($0.50$), with 95\% confidence intervals including $0.50$. Expert 2 achieved a higher accuracy ($0.65$); however, the overall near-chance performance and substantial inter-observer variability indicate that the generated ECGs exhibit realistic morphology and are difficult to distinguish from real clinical ECGs.

\subsubsection{Diagnostic Utility of Synthesized ECGs}
In the second task, the same five cardiologists reviewed
100 segments presented in randomized order, including 50 AF and 50 non-AF
segments, under three conditions: PPG alone, PPG with real ECG, and PPG
with synthesized ECG. Cardiologists could skip any segment they were unable
to interpret. For each cardiologist in each condition, we calculated accuracy
as the proportion of correctly classified segments among those for which that
cardiologist provided a judgment. As reported in Table~\ref{tab:cardiologist_af}, mean accuracy was
0.87 with PPG and synthesized ECG, compared with 0.81 with PPG alone and
0.89 with PPG and real ECG. The synthesized-ECG condition yielded higher
mean sensitivity (0.92 vs.\ 0.88) and lower mean specificity (0.88 vs.\ 0.93)
than the real-ECG condition.

\begin{table}[t]
\centering
\caption{Cardiologist AF diagnosis under different input modalities. Values are reported as mean $\pm$ standard deviation over five cardiologists. Best results are highlighted in bold.}
\label{tab:cardiologist_af}
\small
\setlength{\tabcolsep}{1.5pt}
\renewcommand{\arraystretch}{1.05}
\begin{tabular}{@{}lccc@{}}
\toprule
\textbf{Input Modality} 
& \textbf{Acc.}$\uparrow$ 
& \textbf{Sens.}$\uparrow$ 
& \textbf{Spec.}$\uparrow$ \\
\midrule
PPG only 
& $0.81 \pm 0.13$ 
& $0.68 \pm 0.15$ 
& $0.85 \pm 0.07$ \\

PPG + Real ECG 
& $\mathbf{0.89 \pm 0.07}$ 
& $0.88 \pm 0.09$ 
& $\mathbf{0.93 \pm 0.05}$ \\

PPG + Synth. ECG (Ours) 
& $0.87 \pm 0.09$ 
& $\mathbf{0.92 \pm 0.04}$ 
& $0.88 \pm 0.11$ \\
\bottomrule
\end{tabular}
\end{table}

\section{Conclusion}
In this paper, we propose PPGFlowECG, a two-stage framework that aligns PPG and ECG in a shared latent space via the CardioAlign Encoder and synthesizes ECGs using latent rectified flow. By mitigating cross-modal semantic misalignment and the challenges of modeling high-dimensional waveform dynamics, PPGFlowECG achieves state-of-the-art performance in both reconstruction quality and downstream diagnostic tasks across multiple clinical settings. These results support wearable-first cardiovascular screening and early detection when standard ECG acquisition is unavailable and only PPG is accessible. 

\section*{Acknowledgements}
This work is supported by the Beijing Nova Program (202604841257). The authors thank the cardiologists who participated in the expert evaluation: Guanyu Mu (The Second Hospital of Tianjin Medical University, Tianjin, China), Qinghao Zhao (Peking University People's Hospital, Beijing, China), Xingliang Wu (Tianjin Institute of Cardiology, Tianjin, China), Xinxin Di (The First Affiliated Hospital of USTC, Hefei, China), and Jing Zhao (The First Affiliated Hospital of Anhui Medical University, Hefei, China).

\bibliographystyle{IEEEbib}
\bibliography{icdm2026references}

\appendix







\subsection{Algorithm of PPGFlowECG}
\label{appendix_2}

\begin{algorithm}[h]
\caption{Training in Stage 1: CardioAlign Encoder}
\label{alg:stage1_simple}
\begin{algorithmic}[1]
\State 
\textbf{Input:} Dataset $\mathcal{D}$ of paired samples $(x_{\text{ppg}}, x_{\text{ecg}})$.
\State \textbf{Ensure:} Optimized parameters $\theta_{E_{\text{CA}}}, \theta_{D_{\text{PPG}}}, \theta_{D_{\text{ECG}}}$.
\While{not converged}
    \State Sample a mini-batch $(x_{\text{ppg}}, x_{\text{ecg}}) \sim \mathcal{D}$.
    \For{$m \in \{\text{ppg, ecg}\}$} 
        \State $(\boldsymbol{\mu}_m, \boldsymbol{\sigma}_m) \leftarrow E_{CA}(x_m)$
        \State $\mathbf{z}_m \leftarrow \boldsymbol{\mu}_m + \boldsymbol{\sigma}_m \odot \boldsymbol{\epsilon}$ \quad where $\boldsymbol{\epsilon} \sim \mathcal{N}(\mathbf{0}, \mathbf{I})$
    \EndFor
    \State Compute $\mathcal{L}_{\text{rec}}$, $\mathcal{L}_{\text{kl}}$, $\mathcal{L}_{\text{GDA}}$, $\mathcal{L}_{\text{LID}}$, $\mathcal{L}_{\text{SDC}}$.
    \State $\mathcal{L}_{\text{CA}}$ integrates $\mathcal{L}_{\text{GDA}}$, $\mathcal{L}_{\text{LID}}$, and $\mathcal{L}_{\text{SDC}}$.
    \State$ \mathcal{L}_{\text{stage1}} \leftarrow \sum_{m \in \{\text{ppg,ecg}\}} \left( \mathcal{L}_{\text{rec}}^{(m)} + \alpha\mathcal{L}_{\text{kl}}^{(m)} \right) + \mathcal{L}_{\text{CA}}$ \quad where $\alpha$ is set to $10^{-4}$
    \State Update $\theta_{E_{\text{CA}}}, \theta_{D_{\text{PPG}}}, \theta_{D_{\text{ECG}}}$ using $\nabla \mathcal{L}_{\text{stage1}}$
\EndWhile
\end{algorithmic}
\end{algorithm}

\begin{algorithm}[h]
\caption{Training in Stage~2: Latent Rectified Flow}
\label{alg:stage2}
\begin{algorithmic}[1]
\State \textbf{Input:} Dataset $\mathcal{D}$, frozen encoder $E_{\text{CA}}$, trainable flow model $v_\theta$.
\State \textbf{Ensure:} Optimized parameters $\theta$ for the flow model $v_\theta$.
\While{not converged}
    \State Sample a mini-batch $(x_{\text{ppg}}, x_{\text{ecg}}) \sim \mathcal{D}$.
    \State $c \leftarrow E_{\text{CA}}(x_{\text{ppg}})$ 
    \State $y \leftarrow E_{\text{CA}}(x_{\text{ecg}})$ 
    \State $x_t \leftarrow (1-t)z + ty$ \quad where $z \sim \mathcal{N}(\mathbf{0}, \mathbf{I})$, $t \sim \mathcal{U}(\mathbf{0}, \mathbf{1})$
    \State $\mathcal{L}_{\text{stage2}} \leftarrow \| v_\theta(x_t, t, c) - (y-z) \|_2^2$
    \State Update $\theta$ using $\nabla \mathcal{L}_{\text{stage2}}$
\EndWhile
\end{algorithmic}
\end{algorithm}


\subsection{Implementation Details}
\noindent\textit{Stage~1.} The encoder uses a residual 1D CNN with
self-attention and five $2\times$ temporal downsampling layers. It outputs
$\mu,\sigma\in\mathbb{R}^{L/32\times4}$ for reparameterization and scales the
sampled latent by 0.18215. Each decoder combines residual upsampling
with a TimeVAE-inspired decomposable branch~\cite{desai2021timevae}.
We train with AdamW for 40,000 iterations (batch size 128,
learning rate $2\times10^{-5}$).

\noindent\textit{Stage~2.} The frozen encoder supplies the PPG
condition and ECG target latent. We train the flow with Adam (learning rate
$10^{-4}$, gradient clipping at 1.0) and maintain a parameter EMA (decay 0.995,
updated every 10 steps).

\begin{table}[t]
\centering
\small
\setlength{\tabcolsep}{3.0pt}
\renewcommand{\arraystretch}{1.05}
\caption{Progressive ablation of CardioAlign Encoder.}
\label{tab:progressive_cardioalign_ablation}
\begin{tabular}{@{}lccc cc@{}}
\toprule
\textbf{Variant} 
& \textbf{GDA} 
& \textbf{LID} 
& \textbf{SDC} 
& \textbf{$\text{MAE}_{\text{HR}}$} 
& \textbf{Macro-AUROC} \\
\midrule
Baseline 
& -- & -- & -- 
& 3.28 & 53.1 \\

+ GDA 
& \checkmark & -- & -- 
& 2.11 & 58.3 \\

+ GDA + LID 
& \checkmark & \checkmark & -- 
& 1.97 & 60.7 \\

\midrule
CardioAlign Encoder 
& \checkmark & \checkmark & \checkmark 
& \textbf{1.93} & \textbf{61.5} \\
\bottomrule
\end{tabular}
\end{table}

\subsection{Progressive Ablation of the CardioAlign Encoder}
The baseline uses a direct PPG-to-ECG encoder--decoder trained
with waveform reconstruction alone. On MCMED, adding GDA, LID, and SDC in
sequence lowers heart-rate MAE and increases Macro-AUROC, with the largest
gain after adding GDA (Table~\ref{tab:progressive_cardioalign_ablation}).

\subsection{Visualization of Cross-Modal Latent Alignment}
\label{sec:supp_tsne}

\begin{figure}[t]
    \centering
    \includegraphics[width=0.98\linewidth]{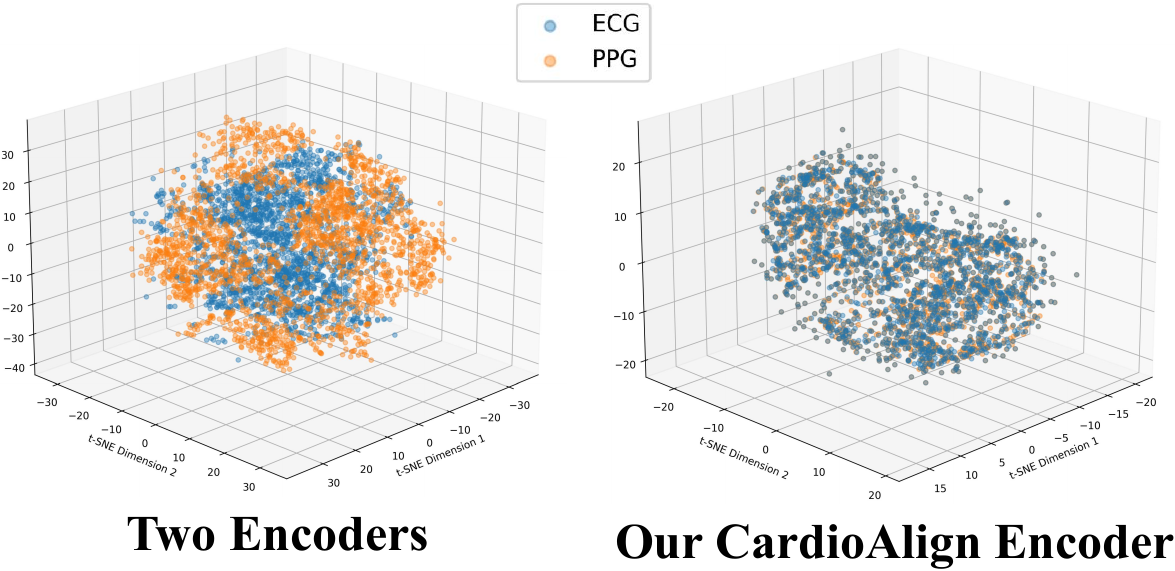}
    \caption{Three-dimensional t-SNE projections of PPG and ECG
    latent representations from two modality-specific encoders (left) and
    the shared CardioAlign Encoder (right). Points are colored by input modality.}
    \label{fig:supp_tsne}
\end{figure}

We use three-dimensional t-SNE~\cite{maaten2008visualizing}
to visualize PPG and ECG latent representations from two encoder
configurations. As shown in Fig.~\ref{fig:supp_tsne}, the modalities are
more visibly separated with two modality-specific encoders. With the shared
CardioAlign Encoder, PPG and ECG points are more intermixed in the projection,
suggesting reduced modality separation in this visualization.

\end{document}